\documentclass[letterpaper,conference, 10pt]{ieeeconf}
\usepackage[T1]{fontenc}
\usepackage[utf8]{inputenc}
\usepackage{units}
\usepackage{amsmath}
\usepackage{amssymb}

\makeatletter

\pdfpageheight\paperheight
\pdfpagewidth\paperwidth

\newtheorem{remrk}{Remark}

\makeatletter
\IEEEoverridecommandlockouts                              % This command is only needed if 
\usepackage{cite}
\usepackage{color}
\usepackage{graphicx}
\usepackage{subfigure}
\makeatother

\makeatother

\begin{document}
\global\long\def\mymatrix#1{\boldsymbol{#1}}

\global\long\def\quat#1{\boldsymbol{#1}}
\global\long\def\dq#1{\underline{\boldsymbol{#1}}}
\global\long\def\hp{\mathbb{H}_{p}}
\global\long\def\dotmul#1#2{\langle#1,#2\rangle}
\global\long\def\partialfrac#1#2{\frac{\partial\left(#1\right)}{\partial#2}}
\global\long\def\totalderivative#1#2{\frac{d\left(#1\right)}{d#2}}
\global\long\def\vecfour#1{\operatorname{vec}_{4}#1}
\global\long\def\haminuseight#1{\overset{-}{\mymatrix H}_{8}\left(#1\right)}
\global\long\def\hapluseight#1{\overset{+}{\mymatrix H}_{8}\left(#1\right)}
\global\long\def\haminus#1{\overset{-}{\mymatrix H}_{4}\left(#1\right)}
\global\long\def\haplus#1{\overset{+}{\mymatrix H}_{4}\left(#1\right)}
\global\long\def\norm#1{\left\Vert #1\right\Vert }
\global\long\def\abs#1{\left|#1\right|}
\global\long\def\conj#1{#1^{*}}
\global\long\def\veceight#1{\operatorname{vec}_{8}#1}
\global\long\def\re#1{Re\left(#1\right)}

\global\long\def\myvec#1{\boldsymbol{#1}}

\global\long\def\imi{\hat{\imath}}

\global\long\def\imj{\hat{\jmath}}

\global\long\def\imk{\hat{k}}

\global\long\def\getp#1{\operatorname{\mathcal{P}}\left(#1\right)}

\global\long\def\getd#1{\operatorname{\mathcal{D}}\left(#1\right)}

\global\long\def\real#1{\operatorname{\mathrm{Re}}\left(#1\right)}

\global\long\def\imag#1{\operatorname{\mathrm{Im}}\left(#1\right)}

\global\long\def\spin{\text{Spin}(3)}

\global\long\def\spinr{\text{Spin}(3){\ltimes}\mathbb{R}^{3}}

\global\long\def\crossmatrix#1{\overline{\mymatrix S}\left(#1\right)}

\global\long\def\getpdot#1{\operatorname{\dot{\mathcal{P}}}\left(#1\right)}

\global\long\def\getddot#1{\operatorname{\dot{\mathcal{D}}}\left(#1\right)}

\global\long\def\squaredistance#1#2#3{d_{#1,#2}^{#3}}

\global\long\def\dotsquaredistance#1#2#3{\dot{d}_{#1,#2}^{#3}}

\global\long\def\dual{\varepsilon}

\title{Active Constraints using Vector Field Inequalities for Surgical Robots}

\author{Murilo M. Marinho, \emph{Student Member, IEEE,} Bruno V. Adorno,
\emph{Senior Member, IEEE,}\\ Kanako Harada, \emph{Member}, \emph{IEEE},
and Mamoru Mitsuishi, \emph{Member}, \emph{IEEE}\thanks{This work
was funded by the ImPACT Program of the Council for Science, Technology
and Innovation (Cabinet Office, Government of Japan). Murilo M. Marinho
has been supported by the Japanese Ministry of Education, Culture,
Sports, Science, and Technology (MEXT). Bruno V. Adorno has been supported
by the Brazilian agencies CAPES, CNPq, FAPEMIG, and by the project
INCT (National Institute of Science and Technology) under the grant
CNPq (Brazilian National Research Council) 465755/2014-3, FAPESP (São
Paulo Research Foundation)2014/50851-0. }\thanks{Murilo M. Marinho,
Kanako Harada, and Mitsuishi Mamoru are with Department of Mechanical
Engineering, the University of Tokyo, Tokyo, Japan. \texttt{Email:\{murilo,
kanako, mamoru\}@nml.t.u-tokyo.ac.jp.} Bruno V. Adorno is with the
Department of Electrical Engineering and with the Graduate Program
in Electrical Engineering - Federal University of Minas Gerais, Belo
Horizonte-MG, Brazil. \texttt{Email: adorno@ufmg.br}.}}
\maketitle
\begin{abstract}
Robotic assistance allows surgeons to perform dexterous and tremor-free
procedures, but is still underrepresented in deep brain neurosurgery
and endonasal surgery where the workspace is constrained. In these
conditions, the vision of surgeons is restricted to areas near the
surgical tool tips, which increases the risk of unexpected collisions
between the shafts of the instruments and their surroundings, in particular
in areas outside the surgical field-of-view. Active constraints can
be used to prevent the tools from entering restricted zones and thus
avoid collisions. In this paper, a vector field inequality is proposed
that guarantees that tools do not enter restricted zones. Moreover,
in contrast with early techniques, the proposed method limits the
tool approach velocity in the direction of the forbidden zone boundary,
guaranteeing a smooth behavior and that tangential velocities will
not be disturbed. The proposed method is evaluated in simulations
featuring two eight degrees-of-freedom manipulators that were custom-designed
for deep neurosurgery. The results show that both manipulator-manipulator
and manipulator-boundary collisions can be avoided using the vector
field inequalities.
\end{abstract}

\section{Introduction}

Surgical robots have received considerable attention in the context
of minimally invasive surgery, as aids in procedures performed through
small incisions, in systems such as the da Vinci, the RAVEN \cite{lum2009raven},
and the SteadyHand \cite{taylor1999steady}. Their use has been extended
to procedures in restricted workspaces such as deep brain microsurgery
\cite{morita2005microsurgical,UEDA2017110} and endonasal surgery
\cite{li2007spatial}. 

In minimally invasive surgery and microsurgery, the surgeon operates
with long thin tools, and views the workspace through an endoscope
or microscope. Their vision is frequently limited to a region near
the surgical tool tips. As the surgeon operates the tools, the tool
shaft may inadvertently collide with neighboring structures. In particular
in microsurgery, the amplitude of hand tremor can be higher than size
of the structures being treated. In this context, surgical robots
are used as equipment to assist the surgeon in order to increase accuracy
and safety, and reduce surgeon's mental load and the effects of their
hand tremor. 

To increase accuracy and attenuate hand tremor, surgical robots are
commanded in task space coordinates either through teleoperation or
comanipulation. Joint space control inputs are generally obtained
by using a kinematic control law, which is usually based on the robot
differential kinematics \cite{Siciliano2009}. Kinematic control laws
are valid when low accelerations are imposed in the joint space, and
are ubiquitous in control algorithms designed for surgical robotics
\cite{kapoor2006constrained,li2007spatial,aghakhani2013task,pham2015analysis,marinho2016using},
since low accelerations are expected in such scenarios. 

An increase in safety and a reduction in the surgeon's mental load
has been achieved through the generation of active constraints (virtual
fixtures) that can, for instance, act as a layer of safety to prevent
the surgical tool from entering a restricted region, even if this
contradicts the commands given to the robot \cite{li2007spatial,kapoor2006constrained},
or transparently generate constrained motion \cite{marinho2016using,pham2015analysis,aghakhani2013task}.
An in-depth survey of active constraints was presented by Bowyer \emph{et
al}. \cite{bowyer2014active}. More recent papers published after
this survey addressed the use of guidance virtual fixtures to assist
in knot tying in robotic laparoscopy \cite{chen2016virtual}, and
to allow surgeons to feel the projected force feedback from the distal
end of a tool to the proximal end of a tool in a comanipulation context
\cite{vitrani2017applying}.

The generation of active constraints using kinematic control laws
requires obtaining a Jacobian that relates the constraint to the task
at hand. For instance, for the generation of the remote center of
motion in minimally invasive surgery, some groups \cite{aghakhani2013task,pham2015analysis}
developed Jacobians to maintain a low remote center of motion error.
Aghakhani \emph{et al.} \cite{aghakhani2013task} constrained lateral
motion in the pivoting point by using the pivoting point dynamics,
and Pham \emph{et al. }\cite{pham2015analysis} used the so-called
constrained Jacobian to project the desired end effector velocity
into a constrained velocity in the pivoting point frame. Additional
requirements, such as the desired tool tip pose following and obstacle
avoidance are usually projected in the null space of the constrained
task or stacked to form an augmented Jacobian. The main problem from
which both approaches suffer is the difficulty in adding inequality
constraints, which are useful in the design of virtual fixtures \cite{funda1996constrained}.

\section{Related works}

A framework for manipulator control in surgical robotics that can
take inequalities into account was developed by Funda \emph{et al}.
\cite{funda1996constrained} using quadratic programming. Their framework
was extended by Kapoor \emph{et al}. \cite{kapoor2006constrained},
who developed a library of virtual fixtures, including five task primitives
that can be combined into customized active constraints. Li \emph{et
al.} \cite{li2007spatial} used the extended framework to aid operators
to move a surgical tool in a highly constrained space in sinus surgery
without collisions. In all these studies, non-linear constraints were
used. Solving for non-linear constraints requires an initial guess
and takes longer than solving for linear constraints, and therefore,
the authors also considered the use of linear \emph{approximations},
which reduce computation time but may result in errors \cite{kapoor2006constrained}.
This raises the question of whether the constraints can be written
directly in linear form, avoiding approximations. An additional consideration
is that the closed-loop stability of their approach has never been
formally proven.

Gonçalves \emph{et al.} \cite{gonccalves2016parsimonious} developed
a Lyapunov-stable kinematic controller that takes into account both
equality and inequality constraints, which are linear with respect
to the joint velocities. They experimentally validated their approach
using a humanoid robot. Their framework was extended by Quiroz-Omana
and Adorno \cite{Quiroz-Omana2017} to the control of a mobile manipulator
byadding a unilateral equality constraint that pushes the mobile manipulator
out of a forbidden zone. This, however, was a reactive constraint
and did not impede the mobile robot from entering the restricted zone
in the first place. Such reactive behavior is particularly undesirable
in medical robotic applications, in which the patient may be harmed
if a tools enters a forbidden zone.

It is important to note that, in prior approaches \cite{li2007spatial,kapoor2006constrained,funda1996constrained},
when the tool reaches a restricted zone boundary the obstacle constraint
is suddenly activated, which might cause the robot to show acceleration
peaks. Researchers attempted to address this issue. For instance,
Xia \emph{et al.} \cite{xia2008integrated} reduced the proportional
gain in an admittance control law proportionally to the distance between
the robot and the nearest obstacle. This allowed the system to smoothly
avoid collisions, but also impeded motion tangential to the obstacle
boundary. Prada and Payandeh \cite{prada2009study} proposed adding
a time-varying gain to smooth the attraction force of attractive virtual
fixtures.

\subsection{Statement of contributions}

In this paper, we propose a new concept of vector field inequality
applied to active constraints that further extrends the developments
of Gonçalves \emph{et al}. \cite{gonccalves2016parsimonious} and
Quiroz-Omana and Adorno \cite{Quiroz-Omana2017}. The proposed vector
field inequality allows the designer to set a maximum approach speed
in the direction of the restriction boundaries, which guarantees that
the robot can smoothly approach restricted zones without trespassing
on their limits, and that the velocities orthogonal to the restriction
boundaries are undisturbed. 

Moreover, in this study we also developed a set of control primitives
for restricted zones, including point, plane, line, and cylinder relations,
which can be customized to model complex environments and interactions.
The effectiveness of the proposed approach is shown in realistic simulated
experiments.

The current work was conducted in the context of \emph{Project 2:
Smart Arm}, part of the\emph{ ImPACT} \emph{Bionic Humanoids Propelling
New Industrial Revolution} project, the objective of which is to develop
robotic systems and control frameworks to allow robot-aided surgical
procedures in constrained spaces, such as those involved in transnasal
pituitary gland resection and surgeries in deep and narrow regions
of the brain. 

\section{Mathematical Background}

The proposed virtual fixtures framework extensively uses dual quaternion
algebra because of its several advantages over other representations.
For instance, unit dual quaternions do not have representational singularities
and are more compact and computationally efficient than homogeneous
transformation matrices \cite{Adorno2011e}. Moreover, their strong
algebraic properties allow different robots to be systematically modeled
\cite{yang1963application,Adorno2011e,Perez2004,Quiroz-Omana2017}
and, in addition to representing rigid motions, dual quaternion algebra
is very useful for describing twists, wrenches, and several geometrical
primitives—e.g., Plücker lines and planes—in very straightforwardly
\cite{adorno2017robot}. The next subsection introduces the basic
definitions of quaternions and dual quaternions; more information
can be found in \cite{yang1963application,adorno2017robot,Selig2005}.

\subsection{Quaternions and dual Quaternions}

Quaternions can be regarded as an extension of complex numbers. The
quaternion set is defined as
\[
\mathbb{H}\triangleq\left\{ h_{1}+\imi h_{2}+\imj h_{3}+\imk h_{4}\,:\,h_{1},h_{2},h_{3},h_{4}\in\mathbb{R}\right\} ,
\]
in which the imaginary units $\imi$, $\imj$, and $\imk$ have the
properties $\hat{\imath}^{2}=\hat{\jmath}^{2}=\hat{k}^{2}=\hat{\imath}\hat{\jmath}\hat{k}=-1$.
The dual quaternion set is defined as
\[
\mathcal{H}\triangleq\left\{ \quat h+\dual\quat h'\,:\,\quat h,\quat h'\in\mathbb{H},\,\dual^{2}=0,\,\dual\neq0\right\} ,
\]
where $\dual$ is the dual (or Clifford) unit \cite{Selig2005}. Addition
and multiplication are defined for dual quaternions analogously to
complex numbers, and hence we need only to respect the properties
of the imaginary and dual units. 

Given $\dq h\in\mathcal{H}$ such that $\dq h=h_{1}+\imi h_{2}+\imj h_{3}+\imk h_{4}+\dual\left(h_{1}'+\imi h_{2}'+\imj h_{3}'+\imk h_{4}'\right)$,
we define the operators
\begin{align*}
\getp{\dq h} & \triangleq h_{1}{+}\imi h_{2}{+}\imj h_{3}{+}\imk h_{4}, & \getd{\dq h} & \triangleq h_{1}'{+}\imi h_{2}'{+}\imj h_{3}'{+}\imk h_{4}',
\end{align*}
and 
\[
\begin{split}\real{\dq h} & \triangleq h_{1}+\dual h_{1}',\\
\imag{\dq h} & \triangleq\imi h_{2}+\imj h_{3}+\imk h_{4}+\dual\left(\imi h_{2}'+\imj h_{3}'+\imk h_{4}'\right).
\end{split}
\]
The conjugate of $\dq h$ is defined as $\conj{\dq h}\triangleq\real{\dq h}-\imag{\dq h}$,
and its norm is given by $\norm{\dq h}=\sqrt{\dq h\conj{\dq h}}=\sqrt{\conj{\dq h}\dq h}$. 

The set $\mathbb{H}_{p}\triangleq\left\{ \quat h\in\mathbb{H}\,:\,\real{\quat h}=0\right\} $
is isomorphic to $\mathbb{R}^{3}$ under the addition operation. Thus,
the quaternion $\left(x\hat{\imath}+y\hat{\jmath}+z\hat{k}\right)\in\mathbb{H}_{p}$
represents the point $\left(x,y,z\right)\in\mathbb{R}^{3}$. The set
of quaternions with unit norm is defined as $\mathbb{S}^{3}\triangleq\left\{ \quat h\in\mathbb{H}\,:\,\norm{\quat h}=1\right\} $,
and $\quat r\in\mathbb{S}^{3}$ can always be written as $\quat r=\cos\left(\phi/2\right)+\quat v\sin\left(\phi/2\right)$,
where $\phi\in\mathbb{R}$ is the rotation angle around the rotation
axis $\quat v\in\mathbb{S}^{3}\cap\mathbb{H}_{p}$ \cite{adorno2017robot}.
Elements of the set $\dq{\mathcal{S}}\triangleq\left\{ \dq h\in\mathcal{H}\,:\,\norm{\dq h}=1\right\} $
are called unit dual quaternions and represent tridimensional poses
(i.e., combined position and orientation) of rigid bodies. Given $\dq x\in\dq{\mathcal{S}}$,
it can always be written as $\dq x=\quat r+\dual\left(1/2\right)\quat t\quat r$,
where $\quat r\in\mathbb{S}^{3}$ and $\quat t\in\mathbb{H}_{p}$
represent the orientation and position, respectively \cite{Selig2005}.
The set $\dq{\mathcal{S}}$ equipped with the multiplication operation
forms the group $\spinr$, which double covers $\mathrm{SE}\left(3\right)$. 

Elements of the set $\mathcal{H}_{p}\triangleq\left\{ \dq h\in\mathcal{H}\,:\,\real{\dq h}=0\right\} $
are called pure dual quaternions and are useful for representing Plücker
lines. More specifically, a Plücker line belongs to the set $\mathcal{H}_{p}\cap\dq{\mathcal{S}}$
and thus is represented by a pure unit dual quaternion such as \cite{adorno2017robot,yang1963application}
\begin{equation}
\dq l=\quat l+\dual\quat m,\label{eq:plucker_line}
\end{equation}
where $\quat l\in\mathbb{H}_{p}\cap\mathbb{S}^{3}$ is a pure quaternion
with unit norm that represents the line direction, and the line moment
is given by $\quat m=\quat p_{l}\times\quat l$, in which $\quat p_{l}\in\mathbb{H}_{p}$
is a point in the line and $\times$ is the cross product. Given $\dq a,\dq b\in\mathcal{H}_{p}$,
the inner product and the cross product are respectively defined as
\cite{yang1963application,adorno2017robot}
\begin{align}
\dotmul{\dq a}{\dq b} & \triangleq-\frac{\dq a\dq b+\dq b\dq a}{2}, & \dq a\times\dq b & \triangleq\frac{\dq a\dq b-\dq b\dq a}{2}.\label{eq:dot and cross products}
\end{align}

The operator $\vecfour{}$ maps quaternions into $\mathbb{R}^{4}$
and $\veceight{}$ maps dual quaternions into $\mathbb{R}^{8}$. For
instance, $\vecfour{\quat h}=\begin{bmatrix}h_{1} & h_{2} & h_{3} & h_{4}\end{bmatrix}^{T}$
and $\veceight{\dq h}=\begin{bmatrix}h_{1} & h_{2} & h_{3} & h_{4} & h_{1}^{'} & h_{2}^{'} & h_{3}^{'} & h_{4}^{'}\end{bmatrix}^{T}$. 

Finally, given $\quat h_{1},\quat h_{2}\in\mathbb{H}$, the Hamilton
operators are matrices that satisfy $\vecfour{\left(\quat h_{1}\quat h_{2}\right)}=\haplus{\quat h_{1}}\vecfour{\quat h_{2}}=\haminus{\quat h_{2}}\vecfour{\quat h_{1}}$.
Analogously, given $\dq h_{1},\dq h_{2}\in\mathcal{H}$, the Hamilton
operators satisfy $\veceight{\left(\dq h_{1}\dq h_{2}\right)}=\hapluseight{\dq h_{1}}\veceight{\dq h_{2}}=\haminuseight{\dq h_{2}}\veceight{\dq h_{1}}$
\cite{adorno2017robot}. 

It can be shown by direct calculation that, when $\quat a,\quat b\in\mathbb{H}_{p}$,
the inner product can be written as
\[
\dotmul{\quat a}{\quat b}=\left(\vecfour{\quat a}\right)^{T}\vecfour{\quat b}=\left(\vecfour{\quat b}\right)^{T}\vecfour{\quat a}
\]
and the cross product between $\quat a$ and $\quat b$, which is
analogous to the case of vectors in $\mathbb{R}^{3}$, can be mapped
into $\mathbb{R}^{4}$ by using the operator $\crossmatrix{\cdot}$
as 
\begin{align*}
\vecfour{\left(\quat a\times\quat b\right)} & =\underbrace{\begin{bmatrix}0 & 0 & 0 & 0\\
0 & 0 & -a_{4} & a_{3}\\
0 & a_{4} & 0 & -a_{2}\\
0 & -a_{3} & a_{2} & 0
\end{bmatrix}}_{\crossmatrix{\quat a}}\vecfour{\quat b}\\
 & =\crossmatrix{\quat a}\vecfour{\quat b}=\crossmatrix{\quat b}^{T}\vecfour{\quat a}.
\end{align*}
Finally, the time-derivative of a quaternion $\quat a\in\mathbb{H}_{p}$
is given by{\small
\begin{align}
\totalderivative{\norm{\quat a}}t=\frac{\dot{\quat a}\conj{\quat a}+\quat a\conj{\dot{\quat a}}}{2\norm{\quat a}}= & \frac{\dotmul{\dot{\quat a}}{\quat a}}{\norm{\quat a}}=\frac{1}{\norm{\quat a}}\vecfour{\quat a^{T}}\vecfour{\dot{\quat a}},\label{eq:quaternion-norm-derivative-1}
\end{align}
}since for any $\quat a\in\mathbb{H}_{p}$ we have $\conj{\quat a}=-\quat a$.

\subsection{Differential kinematics}

Differential kinematics is the relation between task space velocities
and joint space velocities, in the general form $\dot{\myvec x}=\mymatrix J\dot{\myvec q}$,
in which $\dot{\myvec x}$ $\in$ $\mathbb{R}^{m}$ is the vector
of task space velocities, $\dot{\myvec q}$ $\in$ $\mathbb{R}^{n}$
is the vector of manipulator joint velocities, and $\mymatrix J$
$\in$ $\mathbb{R}^{m\times n}$ is the Jacobian matrix.

For instance, given the robot end effector pose $\dq x\triangleq\dq x\left(\myvec q\left(t\right)\right)\in\spinr$,
the differential kinematics is given by $\veceight{\dot{\dq x}}=\mymatrix J_{\dq x}\dot{\myvec q}$,
where $\mymatrix J_{\dq x}$ $\in$ $\mathbb{R}^{8\times n}$ is the
dual quaternion analytical Jacobian, which can be found by using dual
quaternion algebra \cite{Adorno2011e}. Similarly, given the end effector
position $\quat t\triangleq\quat t\left(\quat q\left(t\right)\right)\in\mathbb{H}_{p}$
and the orientation $\quat r\triangleq\quat r\left(\myvec q\left(t\right)\right)\in\mathbb{S}^{3}$,
we have
\begin{align}
\vecfour{\dot{\quat t}} & =\mymatrix J_{t}\dot{\myvec q},\label{eq:translation_jacobian}\\
\vecfour{\dot{\quat r}} & =\mymatrix J_{r}\dot{\myvec q},\label{eq:rotation_jacobian}
\end{align}
where $\mymatrix J_{t},\mymatrix J_{r}\in\mathbb{R}^{4\times n}$
are calculated from $\mymatrix J_{\dq x}$ also by using dual quaternion
algebra \cite{adorno2010dual}. It is important to note that, in general
applications, $\dq x$, $\quat t$, and $\quat r$ may refer to any
relevant coordinate system related to the robot, and not only to the
end effector. 

\subsection{Linear programming for differential inverse kinematics }

In closed-loop differential inverse kinematics, first a desired task
space target, $\myvec x_{d}$, and the task error $\myvec{\tilde{x}}=\myvec x-\myvec x_{d}$
are defined. For $\eta$ $\in$ $(0,\infty)$, the minimum-norm analytical
solution of the optimization problem
\begin{align}
\underset{\dot{\myvec q}}{\min} & \norm{\mymatrix J\dot{\myvec q}+\eta\myvec{\tilde{x}}}_{2}^{2}\label{eq:problem_quadratic}
\end{align}
is $\mymatrix J^{\dagger}\eta\myvec{\tilde{x}}=\dot{\myvec q}$, in
which $\mymatrix J^{\dagger}$ is the generalized Moore-Penrose pseudoinverse
of $\mymatrix J$. 

Adding inequality and equality constraints in Problem~\ref{eq:problem_quadratic}
turns it into a quadratic programming problem requiring a numerical
solver \cite{Escande2014}. The 1-norm analogue of Problem \ref{eq:problem_quadratic}
is
\begin{align}
\underset{\dot{\myvec q}}{\min} & \norm{\mymatrix J\dot{\myvec q}+\eta\myvec{\tilde{x}}}_{1}.\label{eq:problem_linear}
\end{align}
A Lyapunov stable solution for Problem \ref{eq:problem_linear} that
considers both equality and inequality constraints is given by the
following linear program in canonical form \cite{gonccalves2016parsimonious},
\begin{align}
\underset{\myvec g}{\min} & \begin{bmatrix}-\myvec 1^{T}\mymatrix J & \myvec 1^{T}\mymatrix J & 2\cdot\myvec 1^{T} & \myvec 0^{T} & 0 & \myvec 0^{T}\end{bmatrix}\myvec g-\eta\myvec 1^{T}\myvec{\tilde{x}}\nonumber \\
\text{subject to} & \begin{bmatrix}\mymatrix J & -\mymatrix J & -\mymatrix I & \mymatrix I & \mymatrix 0 & \mymatrix 0\\
\myvec 1^{T} & \myvec 1^{T} & \myvec 0^{T} & \myvec 0^{T} & 1 & \myvec 0^{T}\\
\mymatrix W & -\mymatrix W & \mymatrix 0 & \mymatrix 0 & \mymatrix 0 & \mymatrix I
\end{bmatrix}\myvec g=\begin{bmatrix}-\eta\myvec{\tilde{x}}\\
\beta\norm{\myvec{\tilde{x}}}_{1}\\
\myvec w
\end{bmatrix},\label{eq:problem_linear_final}\\
 & \myvec g\geq0\nonumber 
\end{align}
where $\myvec g=\begin{bmatrix}\dot{\myvec q}_{P}^{T} & \dot{\myvec q}_{N}^{T} & \myvec y^{T} & z_{A}^{T} & z_{B} & z_{C}^{T}\end{bmatrix}^{T}$
is the vector of decision variables. Moreover, $\myvec 0^{T}$ and
$\myvec 1^{T}$ are row vectors of ones and zeros of appropriate dimensions,
respectively. In the canonical form, all decision variables should
be non-negative, and hence the joint velocities are split into the
non-negative $\dot{\myvec q}_{P}$ and $\dot{\myvec q}_{N}$ such
that
\begin{equation}
\dot{\myvec q}=\dot{\myvec q}_{P}-\dot{\myvec q}_{N}.\label{eq:control_input_decomposed}
\end{equation}
Moreover, $\myvec y$ is a vector of residuals for the objective function,
and the slack variables $z_{A}$, $z_{B}$, and $z_{C}$ are used
to transform inequality constraints into equality ones. 

The matrix of constraints has three horizontal blocks from top to
bottom. The first block concerns error convergence, the second block
guarantees that the joints will stop moving when $\myvec{\tilde{x}}\rightarrow0$
with $\beta$ $\in\left(0,\infty\right)$, and the third block is
a generic block of $r$ constraints, in which $\mymatrix W$ $\in$
$\mathbb{R}^{r\times n}$ and $\myvec w$ $\in$ $\mathbb{R}^{r\times1}$.
The vector field inequalities proposed in this work concern an appropriate
choice of $\mymatrix W$and $\myvec w$ to generate active constraints,
as described in the following section.

\section{Vector field inequality}

\begin{figure}[h]
\centering 

\def\svgwidth{1\columnwidth}
\footnotesize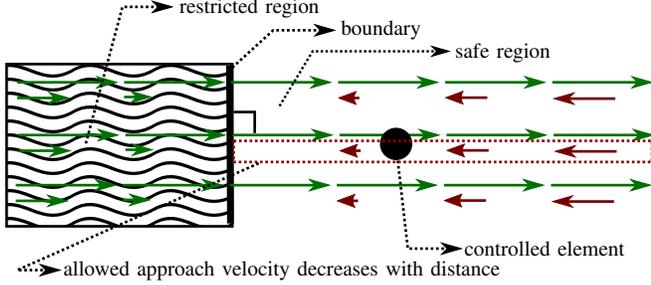

\caption{\label{fig:vectorfieldinequality}Proposed vector field inequality
method. The vector field is given by an inequality constraint, and
to each point in space is assigned a maximum approach velocity (the
red vector in each vector pair), and a maximum separating velocity
(the green vector in each vector pair) in the direction perpendicular
to the restricted zone boundary. Tangential velocities are unconstrained. }
\end{figure}

The vector field inequality proposed in this paper is a general method
for active constraints. It requires:
\begin{enumerate}
\item A function $d(t)$ $\in$ $\mathbb{R}$ that encodes the (signed)
distance between the two collidable entities, and 
\item The Jacobian relating the time derivative of the distance and the
joint velocities in the general form 
\begin{equation}
\dot{d}(t)=\underbrace{\partialfrac{d(t)}{\myvec q}}_{\mymatrix J_{d}}\dot{\myvec q}.\label{eq:distance_kinematics_general}
\end{equation}
\end{enumerate}
Using the distance function and the Jacobian, complex restricted zones
can be generated, by either maintaining the distance above a desired
level or keeping the distance below a certain level. The vector field
inequality is illustrated in Fig. \ref{fig:vectorfieldinequality}.

\subsection{Preventing the robot from entering a restricted region\label{subsec:robot_outside_restricted_region}}

We first define a minimum safe distance, $d_{\text{safe}}$ $\in$
$\left[0,\infty\right)$, that delineates the boundary of the restricted
zone. We then define a distance error as
\begin{equation}
\tilde{d}(t)\triangleq d(t)-d_{\text{safe}},\label{eq:distance_error}
\end{equation}
which will be positive when in the safe-zone, zero along the boundary,
and negative within the restricted zone. 

Assuming a constant safe distance, the distance error dynamics is
given by $\dot{\tilde{d}}(t)=\dot{d}(t).$ A positive $\dot{\tilde{d}}(t)$
means that the system is moving away from the restricted space boundary
and a negative $\dot{\tilde{d}}(t)$ means that the system is moving
closer to the restricted space. The goal is to constrain the distance
dynamics by the inequality
\begin{equation}
\dot{\tilde{d}}(t)\geq-\eta_{d}\tilde{d}(t),\label{eq:lp4_constraint}
\end{equation}
where $\eta_{d}\in\left(0,\infty\right)$. 

To understand Constraint~\ref{eq:lp4_constraint}, let us suppose
that $\tilde{d}(t)>0$, which means that the system is outside the
restricted zone. In this situation, \emph{any }increase in distance
is \emph{always} allowed, which implies $\dot{\tilde{d}}\geq0\geq-\eta_{d}\tilde{d}$.
However, when the system is approaching the boundary from the safe
region (i.e., $0\geq\dot{\tilde{d}}\geq-\eta_{d}\tilde{d}$), it can
approach with a maximum velocity that decreases, in the worst case,
exponentially with the distance $\tilde{d}(t)$, because the maximum
decreasing rate is achieved when $\dot{\tilde{d}}(t)=-\eta_{d}\tilde{d}(t)$.
Thus, the closer the object is to the boundary, the slower it can
move in its direction. Any slower motion toward the boundary is also
allowed, and hence $\dot{\tilde{d}}(t)\geq-\eta_{d}\tilde{d}(t)$.
As soon as $\tilde{d}(t)=0$, the restriction becomes $\dot{\tilde{d}}(t)\geq0$
and therefore the system will not enter the restricted zone.

Considering \eqref{eq:distance_kinematics_general}, Constraint~\ref{eq:lp4_constraint}
is written explicitly in terms of joint velocities as
\begin{alignat}{1}
\mymatrix J_{d}\dot{\myvec q}\geq-\eta_{d}\tilde{d}(t) & \iff-\mymatrix J_{d}\dot{\myvec q}\leq\eta_{d}\tilde{d}(t).\label{lp4_constrained_jacobian}
\end{alignat}
To turn Constraint~\ref{lp4_constrained_jacobian} into an equality
and fit it into the linear programming formalism, we introduce the
slack variable $z_{D}$ and decompose $\dot{\myvec q}$ as in \eqref{eq:control_input_decomposed}
to find
\begin{equation}
-\mymatrix J_{d}\dot{\myvec q}_{P}+\mymatrix J_{d}\dot{\myvec q}_{N}+z_{D}=\eta_{d}\tilde{d}(t).\label{eq:lp4_solution}
\end{equation}
Notice that any number of constraints in the form of Constraint~\ref{eq:lp4_solution}
can be found for different interactions within the robot workspace.
\begin{remrk}
Consider that the system is initially inside the restricted zone;
that is, $\tilde{d}(t)<0$. In this case, Constraint~\ref{eq:lp4_constraint}
will be fulfilled only if $\dot{\tilde{d}}(t)$ is greater than zero,
which means that the system will, at least, be pushed back to the
boundary exponentially. 
\end{remrk}

\subsection{Maintaining the robot inside a safe region}

Using the same reasoning as in Section~\ref{subsec:robot_outside_restricted_region},
if we want to maintain the robot \emph{inside} a safe region, we must
redefine $d_{\text{safe}}$; that is,
\[
\tilde{d}(t)\triangleq d_{\text{safe}}-d(t),
\]
with the final solution, assuming the desired error dynamics \eqref{eq:lp4_constraint},
given by
\[
\mymatrix J_{d}\dot{\myvec q}_{P}-\mymatrix J_{d}\dot{\myvec q}_{N}+z_{D}=\eta_{d}\tilde{d}(t).
\]

\section{Distance functions and Jacobians for active constraints}

The proposed vector field inequality discussed in the previous section
requires a distance function $d(t)$ and the corresponding distance
Jacobian $\mymatrix J_{d}$ for each collidable pair. In this section,
we find four distance functions and Jacobians for pairs in which one
element is static and the second element is dynamic. 

\subsection{Point–static-plane distance Jacobian, $\protect\mymatrix J_{t,\pi}$}

One of the primitives for virtual fixtures is the constraint of a
point such that it is above a static plane. Given a reference frame
$\mathcal{F}$, a plane $\dq{\pi}$ in dual quaternion space is given
by \cite{adorno2017robot}
\[
\dq{\pi}\triangleq\quat n_{\pi}+\dual d_{\pi},
\]
in which $\quat n_{\pi}\in\mathbb{H}_{p}\cap\mathbb{S}^{3}$ is the
normal to the plane and $d_{\pi}$ $\in$ $\mathbb{R}$ is the signed
perpendicular distance between the plane and the origin of the reference
frame. Moreover, the distance $d_{\pi}$ is given by $d_{\pi}=\dotmul{\quat p_{\pi}}{\quat n_{\pi}}$,
where $\quat p_{\pi}$ is an arbitrary point in the plane. If $\mathcal{F}_{\pi}$
is a frame attached to the plane, the signed distance between $\quat t$,
an arbitrary point in the robot kinematic chain, and the plane, from
the point of view of the plane, is given by
\begin{equation}
d_{t,\pi}^{\pi}\triangleq\dotmul{\quat t}{\quat n_{\pi}}-d_{\pi}.\label{eq:point-static-plane-distance}
\end{equation}
The time derivative of \eqref{eq:point-static-plane-distance} is
\begin{alignat}{1}
\totalderivative{d_{t,\pi}^{\pi}}t\underset{\eqref{eq:translation_jacobian}}{=} & \dotmul{\dot{\quat t}}{\quat n_{\pi}}=\underbrace{\left(\vecfour{\quat n_{\pi}}\right)^{T}\mymatrix J_{t}}_{\mymatrix J_{t,\pi}}\dot{\myvec q}.\label{eq:point-static-plane-jacobian}
\end{alignat}

\subsection{Line Jacobian, $\protect\mymatrix J_{l_{z}}$}

The line Jacobian is a pre-requisite for the more complex Jacobians
described in the following sections. 

The Plücker line \cite{adorno2017robot} $\dq l_{z}\in\mathcal{H}_{p}\cap\dq{\mathcal{S}}$
collinear to the $z$-axis of a frame represented by $\dq x=\quat r+\dual\left(1/2\right)\quat t\quat r$
is given by
\begin{equation}
\dq l_{z}=\quat l_{z}+\dual\quat m_{z},\label{eq:line_z}
\end{equation}
where $\quat l_{z}=\quat r\hat{k}\conj{\quat r}$ and $\quat m_{z}=\quat t\times\quat l_{z}.$
The derivative of \eqref{eq:line_z} is 
\begin{align}
\dot{\dq l}_{z} & =\dot{\quat l}_{z}+\dual\dot{\quat m}_{z}.\label{eq:lz_derivative}
\end{align}
Hence,
\begin{align}
\vecfour{\dot{\quat l}_{z}} & \underset{\eqref{eq:rotation_jacobian}}{=}\underbrace{\left(\haminus{\hat{k}\conj{\quat r}}+\haplus{\quat r\hat{k}}\boldsymbol{C}_{4}\right)\mymatrix J_{r}}_{\mymatrix J_{r_{z}}}\dot{\myvec q}\label{eq:line_direction_jacobian}
\end{align}
in which $\boldsymbol{C}_{4}=\text{diag\ensuremath{\left(1,-1,-1,-1\right)}}.$
In addition,

{\small
\begin{gather*}
\vecfour{\dot{\quat m}_{z}}{=}\underbrace{\frac{\left(\haminus{\quat l_{z}}{-}\haplus{\quat l_{z}}\right)\!\mymatrix J_{t}{+}\left(\haplus{\quat t}{-}\haminus{\quat t}\right)\!\mymatrix J_{r_{z}}}{2}}_{\mymatrix J_{m_{z}}}\dot{\myvec q}.
\end{gather*}
}Finally, \eqref{eq:lz_derivative} can be re-written in term of
joint velocities as
\begin{align}
\veceight{\dot{\dq l}_{z}} & =\begin{bmatrix}\mymatrix J_{r_{z}}\\
\mymatrix J_{m_{z}}
\end{bmatrix}\dot{\myvec q}\triangleq\mymatrix J_{l_{z}}\dot{\myvec q}.\label{eq:robot_line_jacobian}
\end{align}

\subsection{Line–static-point distance Jacobian, $\protect\mymatrix J_{l_{z},p}$}

The line–static-point Jacobian can be used to generate a remote center
of motion. First, we notice that the distance between $\dq l_{z}\in\mathcal{H}_{p}\cap\dq{\mathcal{S}}$
and an arbitrary static point $\quat p\in\mathbb{H}_{p}$ is given
by
\begin{equation}
d_{l_{z},p}=\norm{\quat p\times\quat l_{z}-\quat m_{z}}.\label{eq:line-static-point-distance}
\end{equation}
The derivative of \eqref{eq:line-static-point-distance} is

{\footnotesize
\begin{alignat}{1}
\dot{d}_{l_{z},p} & =\frac{1}{d_{l_{z},p}}\vecfour{\left(\quat p\times\quat l_{z}-\quat m_{z}\right)^{T}}\vecfour{\totalderivative{\left(\quat p\times\quat l_{z}-\quat m_{z}\right)}t}\nonumber \\
 & \underset{\eqref{eq:robot_line_jacobian}}{=}\underbrace{\frac{1}{d_{l_{z},p}}\vecfour{\left(\quat p\times\quat l_{z}-\quat m_{z}\right)^{T}}\left(\crossmatrix{\quat p}\mymatrix J_{r_{z}}-\mymatrix J_{m_{z}}\right)}_{\mymatrix J_{l_{z},p}}\dot{\myvec q}.\label{eq:line-static-point-jacobian}
\end{alignat}

}

\subsection{Point–static-line distance Jacobian, $\protect\mymatrix J_{t,l}$}

The point–static-line Jacobian can be used to keep a point inside/outside
a cylinder. First, we notice that the distance between an arbitrary
line $\dq l\in\mathcal{H}_{p}\cap\dq{\mathcal{S}}$, such that $\dq l=\quat l+\dual\quat m$,
and a point $\quat t\in\mathbb{H}_{p}$ in the robot kinematic chain
is given by
\begin{equation}
d_{t,l}=\norm{\quat t\times\quat l-\quat m}.\label{eq:point-static-line-distance}
\end{equation}
The derivative of \eqref{eq:point-static-line-distance} is given
by
\begin{alignat}{1}
\dot{d}_{t,l} & =\frac{1}{d_{t,l}}\vecfour{\left(\quat t\times\quat l-\quat m\right)^{T}}\vecfour{\totalderivative{\left(\quat t\times\quat l-\quat m\right)}t}\nonumber \\
 & \underset{\eqref{eq:translation_jacobian}}{=}\underbrace{\frac{1}{d_{t,l}}\vecfour{\left(\quat t\times\quat l-\quat m\right)^{T}}\crossmatrix{\quat l}^{T}\mymatrix J_{t}}_{\mymatrix J_{t,l}}\dot{\myvec q}.\label{eq:point-static-line-jacobian}
\end{alignat}

\subsection{Line–static-line distance Jacobian}

The line–static-line Jacobian is particularly useful for avoiding
collisions between a moving cylinder and a static cylinder. In order
to obtain the line–static-line Jacobian, we use the concept of dual
angle between Plucker lines. 

\subsubsection{Inner product Jacobian, $\protect\mymatrix J_{\protect\dotmul{\protect\dq l_{z}}{\protect\dq l}}$}

The dual cosine between Plücker lines $\dq l,\dq l_{z}\in\mathcal{H}_{p}\cap\dq{\mathcal{S}}$
is obtained by using the inner product \cite{yang1963application}
\begin{gather}
\dotmul{\dq l}{\dq l_{z}}=\underbrace{\norm{\dq l}}_{1}\underbrace{\norm{\dq l_{z}}}_{1}\cos\left(\phi+\dual d_{l}\right)=\cos\phi-\dual d_{l}\sin\phi,\label{eq:dual_distance}
\end{gather}
where $d_{l}\in[0,\infty)$ and $\phi\in[0,2\pi)$ are the distance
and the angle between the lines, respectively.\footnote{Given a function $f\,:\,\mathbb{D}\to\mathbb{D}$, where $\mathbb{D}\triangleq\left\{ \dq h\in\mathcal{H}\,:\,\imag{\dq h}=0\right\} $,
it is possible to show that $f\left(a+\dual b\right)=f\left(a\right)+\dual bf'\left(a\right)$.
For more details, see \cite{adorno2017robot}.}

Since the line $\dq l$ is static, $\dot{\dq l}=0$ and thus the derivative
of \eqref{eq:dual_distance} is given by
\begin{align}
\veceight{\totalderivative{\dotmul{\dq l_{z}}{\dq l}}t} & =\underbrace{-\frac{\left(\hapluseight{\dq l}+\haminuseight{\dq l}\right)}{2}\mymatrix J_{l}}_{\mymatrix J_{\dotmul{\dq l_{z}}{\dq l}}}\dot{\myvec q}\nonumber \\
\begin{bmatrix}\vecfour{\getpdot{\dotmul{\dq l_{z}}{\dq l}}}\\
\vecfour{\getddot{\dotmul{\dq l_{z}}{\dq l}}}
\end{bmatrix} & =\begin{bmatrix}\mymatrix J_{\getp{\dotmul{\dq l_{z}}{\dq l}}}\\
\mymatrix J_{\getd{\dotmul{\dq l_{z}}{\dq l}}}
\end{bmatrix}\dot{\myvec q}.\label{eq:dotprodjacobian}
\end{align}

\subsubsection{Cross product Jacobian, $\protect\mymatrix J_{\protect\dq l_{z}\times\protect\dq l}$}

Given the dual angle $\dq{\phi}=\phi+\dual d_{l}$, where $d_{l}\in[0,\infty)$
and $\phi\in[0,2\pi)$, and the Plücker line $\dq s\in\mathcal{H}_{p}\cap\dq{\mathcal{S}}$,
with $\dq s=\quat s+\dual\quat m_{s}$, the cross product between
$\dq l_{z}$ and $\dq l$ is given by \cite{yang1963application}
\begin{align}
\dq l_{z}\times\dq l & =\underbrace{\norm{\dq l_{z}}}_{1}\underbrace{\norm{\dq l}}_{1}\dq s\sin\dq{\phi}\nonumber \\
 & =\left(\quat s+\dual\quat m_{s}\right)\left(\sin\phi+\dual d_{l}\cos\phi\right)\nonumber \\
 & =\quat s\sin\phi+\dual\left(\quat m_{s}\sin\phi+\quat sd_{l}\cos\phi\right),\label{eq:line-dual-sine}
\end{align}
in which $\dq s$ is the line perpendicular to both $\dq l_{z}$ and
$\dq l$. The time derivative of the cross product between $\dq l_{z}$
and the static line $\dq l$ is
\begin{align}
\veceight{\totalderivative{\dq l_{z}\times\dq l}t} & =\underbrace{\frac{\left(\haminuseight{\dq l}-\hapluseight{\dq l}\right)}{2}\mymatrix J_{l}}_{\mymatrix J_{\dq l_{z}\times\dq l}}\dot{\myvec q}\nonumber \\
\begin{bmatrix}\vecfour{\getpdot{\dq l_{z}\times\dq l}}\\
\vecfour{\getddot{\dq l_{z}\times\dq l}}
\end{bmatrix} & =\begin{bmatrix}\mymatrix J_{\getp{\dq l_{z}\times\dq l}}\\
\mymatrix J_{\getd{\dq l_{z}\times\dq l}}
\end{bmatrix}\dot{\myvec q}.\label{eq:crossprodjacobian}
\end{align}

\subsubsection{Distance Jacobian between non-parallel lines}

The distance $d_{\quat l_{z},\quat l\not\parallel}$ between $\dq l_{z}$
and $\dq l$ when they are not parallel (i.e., $\phi\in(0,2\pi)\setminus\pi)$
can be obtained as 
\begin{equation}
d_{\quat l_{z},\quat l\not\parallel}=\frac{\norm{\getd{\dotmul{\dq l_{z}}{\dq l}}}}{\norm{\getp{\dq l_{z}\times\dq l}}}=\frac{\norm{d_{l}\sin\phi}}{\norm{\quat s_{l}\sin\phi}}=d_{l}.\label{eq:line-static-line-nonparallel-distance}
\end{equation}

Noting that both the numerator and denominator of \eqref{eq:line-static-line-nonparallel-distance}
are real numbers, we can find the derivative of \eqref{eq:line-static-line-nonparallel-distance}
as
\begin{multline}
\dot{d}_{\quat l_{z},\quat l\not\parallel}=\underbrace{\frac{1}{\norm{\getp{\dq l_{z}\times\dq l}}}}_{a}\totalderivative{\norm{\getd{\dotmul{\dq l_{z}}{\dq l}}}}t\\
\underbrace{-\frac{\norm{\getd{\dotmul{\dq l_{z}}{\dq l}}}}{\norm{\getp{\dq l_{z}\times\dq l}}^{2}}}_{b}\totalderivative{\norm{\getp{\dq l_{z}\times\dq l}}}t.\label{eq:non-parallel-squared-distance-jacobian}
\end{multline}
We obtain the derivative of the norm of $\getd{\dotmul{\dq l_{z}}{\dq l}}$
using \eqref{eq:quaternion-norm-derivative-1} and \eqref{eq:dotprodjacobian}
as

{\footnotesize
\begin{align}
\totalderivative{\norm{\getd{\dotmul{\dq l_{z}}{\dq l}}}}t & =\underbrace{\frac{1}{\norm{\getd{\dotmul{\dq l_{z}}{\dq l}}}}\vecfour{\getd{\dotmul{\dq l_{z}}{\dq l}}^{T}}\mymatrix J_{\getd{\dotmul{\dq l_{z}}{\dq l}}}}_{\mymatrix J_{\norm{\getd{\dotmul{\dq l_{z}}{\dq l}}}}}\dot{\myvec q}\label{eq:nonparalleldual}
\end{align}
}and similarly we obtain the derivative of the norm of $\getp{\dq l_{z}\times\dq l}$
using \eqref{eq:quaternion-norm-derivative-1} and \eqref{eq:crossprodjacobian}
as

{\footnotesize
\begin{align}
\totalderivative{\norm{\getp{\dq l_{z}\times\dq l}}}t & =\underbrace{\frac{1}{\norm{\getp{\dq l_{z}\times\dq l}}}\vecfour{\getp{\dq l_{z}\times\dq l}^{T}}\mymatrix J_{\getp{\dq l_{z}\times\dq l}}}_{\mymatrix J_{\norm{\getp{\dq l_{z}\times\dq l}}}}\dot{\myvec q}.\label{eq:nonparallelprim}
\end{align}
}Finally, we substitute \eqref{eq:nonparalleldual} and \eqref{eq:nonparallelprim}
in \eqref{eq:non-parallel-squared-distance-jacobian} to find
\begin{align}
\dot{d}_{\quat l_{z},\quat l\not\parallel} & =\underbrace{\left(a\mymatrix J_{\norm{\getd{\dotmul{\dq l_{z}}{\dq l}}}}+b\mymatrix J_{\norm{\getp{\dq l_{z}\times\dq l}}}\right)}_{\mymatrix J_{\quat l_{z},\quat l\not\parallel}}\dot{\myvec q}.\label{eq:nonparalleldistancejocobian}
\end{align}

\subsubsection{Parallel distance Jacobian, $\protect\mymatrix J_{\protect\quat l_{z},\protect\quat l\parallel}$\label{subsec:parallel-distance}}

In the degenerate case in which $\dq l_{z}$ and $\dq l$ are parallel
(i.e., $\phi\in\left\{ 0,\pi\right\} $), the distance between them
can be retrieved as
\begin{equation}
\squaredistance{\quat l_{z}}{\quat l\parallel}{}\triangleq\norm{\getd{\dq l_{z}\times\dq l}}=\norm{\quat sd_{l}}=d_{l}.\label{eq:line-static-line-parallel-distance}
\end{equation}
Resorting once more to \eqref{eq:quaternion-norm-derivative-1}, we
find the derivative of \eqref{eq:line-static-line-parallel-distance}
as{\small
\begin{align}
\dotsquaredistance{\quat l_{z}}{\quat l\parallel}{} & =\underbrace{\frac{1}{\norm{\getd{\dq l_{z}\times\dq l}}}\vecfour{\left(\getd{\dq l_{z}\times\dq l}\right)}^{T}\mymatrix J_{\getd{\dq l_{z}\times\dq l}}}_{\mymatrix J_{\quat l_{z},\quat l\parallel}}.\label{eq:paralleldistancejacobian}
\end{align}
}

\subsubsection{Manipulator line-to-line distance Jacobian, $\protect\mymatrix J_{\protect\quat l_{z},\protect\quat l}$\label{subsec:manipulator-line-line-distance}}

The line-to-line distance Jacobian is given by combining \eqref{eq:nonparalleldistancejocobian}
and \eqref{eq:paralleldistancejacobian} as
\begin{alignat}{2}
\mymatrix J_{\quat l_{z},\quat l} & =\begin{cases}
\mymatrix J_{\quat l_{z},\quat l\not\parallel} & \phi\in(0,2\pi)\setminus\pi\\
\mymatrix J_{\quat l_{z},\quat l\parallel} & \phi\in\left\{ 0,\pi\right\} 
\end{cases} & ,\label{eq:manipulator-line-line-distance-jacobian-residual}
\end{alignat}
in which $\phi$ can be obtained as $\phi=\arccos\getp{\dotmul{\dq l}{\dq l_{z}}}$.

The distance between lines can be given by the composition of the
distance \eqref{eq:line-static-line-nonparallel-distance} and \eqref{eq:line-static-line-parallel-distance}
as
\begin{equation}
d_{\quat l_{z},\quat l}=\begin{cases}
d_{\quat l_{z},\quat l\not\parallel} & \phi\in(0,2\pi)\setminus\pi\\
d_{\quat l_{z},\quat l\parallel} & \phi\in\left\{ 0,\pi\right\} 
\end{cases}.\label{eq:line-line-distance}
\end{equation}

\section{Simulations}

\begin{figure}
\centering

\def\svgwidth{1\columnwidth}
\footnotesize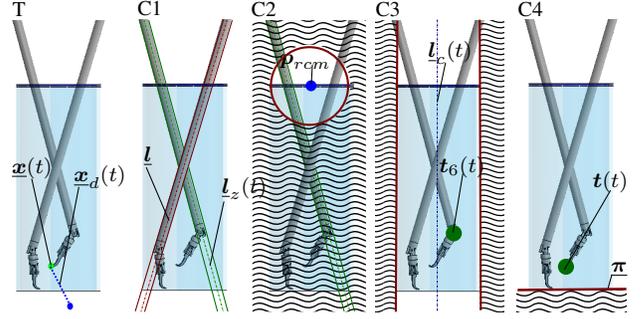

\caption{\label{fig:setup}Task and constraints used in the simulations. The
task T consists of moving the end effector, $\protect\dq x\left(t\right)$,
of one of the tools in a given trajectory $\protect\dq x_{d}(t)$.
The constraints are: C1, a line-static-line distance constraint, to
avoid collisions between tool shafts, where the shaft of the moving
tool has a center-line given by $\protect\dq l_{z}(t)$ and the static
shaft has a center-line $\protect\dq l$; C2, a line-static-point
distance constraint to generate the entry sphere with the center given
by $\protect\quat p_{\text{rcm}}$; C3, a point-static-line distance
constraint to prevent the lower part of the moving tool given by $\protect\quat t_{6}(t)$
from colliding with the cylindrical workspace with center-line $\protect\dq l_{c}$;
and C4, a point-static-plane distance constraint to prevent the end
effector from crossing the lower bound of the cylindrical workspace
given by the plane $\protect\dq{\pi}$.}
\end{figure}

In order to validate our approach,\footnote{See accompanying video.}
we developed a realistic simulated version, using V-REP\footnote{http://www.coppeliarobotics.com/}
and DQ Robotics for MATLAB\footnote{http://dqrobotics.sourceforge.net},
of a deep neurosurgical robotic system that we previously developed
in close cooperation with neurosurgeons \cite{UEDA2017110} (see Fig.~\ref{fig:setup}).
In deep neurosurgery, the workspace is highly constrained in the microscopic
view. The diameter of the tools is $3.5\,\unit{mm}$, and the radius
and depth of the workspace cylinder are $2.8\,\unit{cm}$ and $8\,\unit{cm}$,
respectively.

In this preliminary evaluation, one of the tools was kept fixed while
the second tool moved. Five simulations (S1-S5) with increasing complexity
were run, as follows. S1: the task T consisted of moving the end effector
along the desired trajectory with no virtual fixtures; S2: constraint
C1 iwas added to prevent the shafts from colliding with each other
using $\mymatrix J_{l_{z},l}$ from \eqref{eq:line-line-distance}
and $d_{l_{z},l}$ as in \eqref{eq:line-line-distance}; S3: in addition
to C1, a spherical entry-point constraint C2 was added to avoid collisions
with the top part of the cylinder using $\mymatrix J_{l_{z},p}$ from
\eqref{eq:line-static-point-jacobian} and $d_{l_{z},p}$ as in \eqref{eq:line-static-point-distance};
S4: in addition to constraints C1 and C2, a point line distance constraint
C3 was added to avoid collisions of the lower part of the shaft with
the workspace boundary, which is represented by a cylinder, using
$\mymatrix J_{t,l}$ from \eqref{eq:point-static-line-jacobian} and
$d_{t,l}$ as in \eqref{eq:point-static-line-distance}; and finally,
S5: in addition to constraints C1, C2, and C3, a point–static-plane
distance constraint C4 was added to prevent the end effector from
crossing the plane $\dq{\pi}$, using $\mymatrix J_{t,\pi}$ from
\eqref{eq:point-static-plane-jacobian} and $d_{t,\pi}$ as in \eqref{eq:point-static-plane-distance}.

The robot has eight degrees-of-freedom in which the first, second,
and fifth joints were prismatic and the remaining joints were revolute.
The robots' initial configurations were such that the shafts were
at a long distance from each other. The trajectory, which was the
same for all simulations, was the screwlinear interpolation of four
poses. The final form of the linear program, when all constraints
were used, was

{\small 
\begin{align*}
\underset{\myvec g}{\min} & \left[\begin{array}{ccccc|c|cccc}
-\myvec 1^{T}\mymatrix J_{\dq x} & \myvec 1^{T}\mymatrix J_{\dq x} & 2\cdot\myvec 1^{T} & \myvec 0^{T} & 0 & \myvec 0^{T} & 0 & 0 & 0 & 0\end{array}\right]\myvec g\\
\text{s.t.} & \left[\begin{array}{cccccccccc}
\mymatrix J_{\dq x} & -\mymatrix J_{\dq x} & -\mymatrix I & \mymatrix I & \mymatrix 0 & \mymatrix 0 & \mymatrix 0 & \mymatrix 0 & \mymatrix 0 & 0\\
\myvec 1^{T} & \myvec 1^{T} & \myvec 0^{T} & \myvec 0^{T} & \begin{array}{c}
1\end{array} & \myvec 0^{T} & 0 & 0 & 0 & 0\\
\hline \mymatrix W_{l} & -\mymatrix W_{l} & \mymatrix 0 & \mymatrix 0 & \mymatrix 0 & \mymatrix I & \mymatrix 0 & \mymatrix 0 & \mymatrix 0 & 0\\
\hline \mymatrix J_{l_{z},l} & -\mymatrix J_{l_{z},l} & \myvec 0^{T} & \myvec 0^{T} & 0 & \myvec 0^{T} & 1 & 0 & 0 & 0\\
-\mymatrix J_{l_{z},p} & \mymatrix J_{l_{z},p} & \myvec 0^{T} & \myvec 0^{T} & 0 & \myvec 0^{T} & 0 & 1 & 0 & 0\\
-\mymatrix J_{t,l} & \mymatrix J_{t,l} & \myvec 0^{T} & \myvec 0^{T} & 0 & \myvec 0^{T} & 0 & 0 & 1 & 0\\
\mymatrix J_{t,\pi} & -\mymatrix J_{t,\pi} & \myvec 0^{T} & \myvec 0^{T} & 0 & \myvec 0^{T} & 0 & 0 & 0 & 1
\end{array}\right]\myvec g=\begin{bmatrix}-\eta\veceight{\tilde{\dq x}}\\
\beta\norm{\veceight{\tilde{\dq x}}}_{1}\\
\hline \myvec w_{l}\\
\hline \eta_{l_{z},l}\tilde{d}_{C1}\\
\eta_{l_{z},p}\tilde{d}_{C2}\\
\eta_{t,l}\tilde{d}_{C3}\\
\eta_{t,\pi}\tilde{d}_{C4}
\end{bmatrix}\\
 & \myvec g\geq0\\
 & \myvec g=\begin{bmatrix}\dot{\myvec q}_{P}^{T} & \dot{\myvec q}_{N}^{T} & \myvec y^{T} & z_{A}^{T} & z_{B} & z_{l}^{T} & z_{C1} & z_{C2} & z_{C3} & z_{C4}\end{bmatrix}^{T},
\end{align*}
}in which $\mymatrix W_{l}$ and $\myvec w_{l}$ are constraints
for the joint limits \cite{Quiroz-Omana2017}, $z_{l}^{T},z_{C1},z_{C2},z_{C3},z_{C4}$
are slack variables, $\tilde{d}_{C1}=d_{\text{safe},C1}-d_{l_{z},l}$,
$\tilde{d}_{C2}=d_{l_{z},p}-d_{\text{safe},C2}$, $\tilde{d}_{C3}=d_{t,l}-d_{\text{safe},C3}$,
and $\tilde{d}_{C4}=d_{\text{safe},C4}-d_{t,\pi}$. 

For all simulations, the trajectory tracking gain was $\eta=50$,
and $\beta=40$. Whenever present in a given simulation, vector field
inequality gains were set at the same value $\eta_{l_{z},l}=\eta_{l_{z},p}=\eta_{t,l}=\eta_{t,\pi}=0.5$.
The safety distances for each constraint were $d_{\text{safe},C1}=5\,\unit{mm}$,
$d_{\text{safe},C2}=d_{\text{safe},C3}=14\,\unit{mm}$, and $d_{\text{safe},C4}=0$.

\section{Results and discussion}

\begin{figure*}
\centering 

\def\svgwidth{2.0\columnwidth}
\small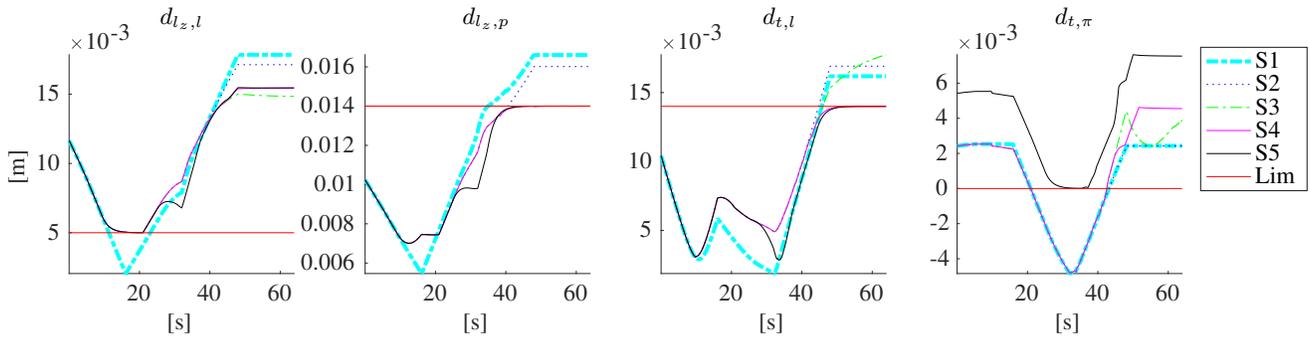

\caption{Distance between collidable entities. Only S5 provided a collision-safe
trajectory, since S4 violated the plane, S3 violated the plane and
the lower part of the cylinder, S2 violated the plane and both the
upper and lower parts of the cylinder, and S1 violated all constraints.
\label{fig:results}}
\end{figure*}

\begin{figure}
\centering 

\def\svgwidth{1.0\columnwidth}
\small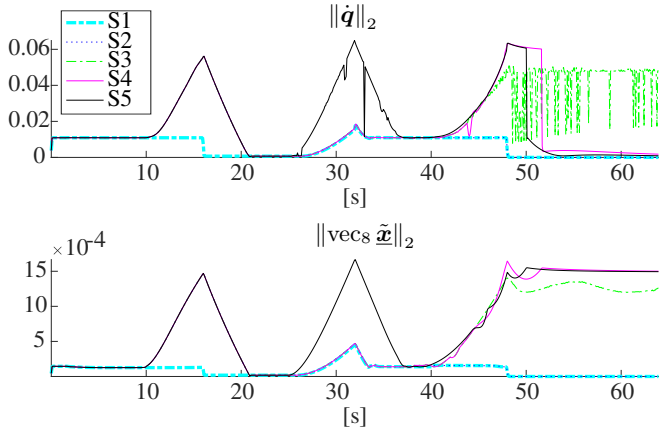

\caption{Trajectory tracking error (\emph{bottom}) and 2-norm of joint velocities
(\emph{top}). Only S1 could closely follow the prescribed trajectory
as it had no active constraints. In all other simulations, some trajectory
tracking error was allowed in order to avoid collisions. \label{fig:results-1}}
\end{figure}

\begin{figure}
\centering 

\def\svgwidth{1.0\columnwidth}
\small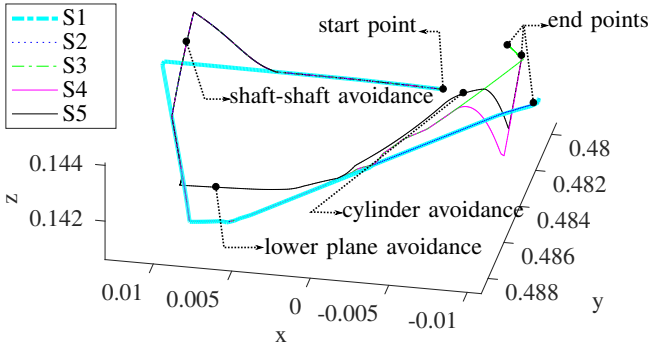

\caption{Tool tip trajectories for all experiments. Notice that S1 is effectively
the same as the desired trajectory, and S5 was the only collision-free
trajectory. \label{fig:results-2}}
\end{figure}

The results of the simulation are shown in Fig.~\ref{fig:results}
in terms of the distance between shafts, the distance between the
entry point and tool shaft, the distance between the lower point of
the shaft and the cylinder centerline, and the distance between the
tool tip and the lower plane.

Simulation S1, in which no virtual fixtures were applied caused collisions
between the shafts and between the moving tool and the workspace boundaries.
Constraining only the shafts in S2 was sufficient to avoid collisions
between the shafts, but tool-boundary collisions still occurred. Constraining
both the shafts and the entry point in S3 provided a collision-free
shaft motion and prevented collisions with the uppermost part of the
cylinder. Simulation S4 violated only the plane constraint, and S5
provided a collision-free path. From the point of view of trajectory
error, as shown in Fig.~\ref{fig:results-1}, naturally S1 was able
to follow the trajectory closely since no active constraints were
applied. All other simulations presented tracking errors given the
gradual imposition of restrictions. The trajectories for all the experiments
are shown in Fig.~\ref{fig:results-2}.

In S3, the robot showed a low amplitude vibration near the end of
the trajectory, which may occur when the trajectory tracking error
is unable to stabilize at the origin. It can be somewhat reduced by
tuning $\beta$; however, other solutions, such as regularizing the
objective function, are being analyzed in ongoing research.

\section{Conclusions}

In this paper a novel method for applying active constraints using
vector field inequalities was proposed. The method can be used to
prevent a robot from entering a restricted zone or for maintaining
its location inside a safe area, and is based on a Lyapunov stable
solution that can handle both equality and inequality constraints.
The vector field inequalities limit the velocity of the robot in the
direction of the forbidden zone's boundary while leaving tangential
velocities undisturbed. To use the method, a Jacobian describing the
relation between the time derivative of the distance and joint velocities
is required. In this work, four of the Jacobians are obtained for
relations in which one of the entities is static. Simulation results
for two eight-DOF neurosurgical arms showed that, by using the vector
field inequalities and relevant primitives, all collisions could be
avoided and the robot safely navigated through a restricted workspace. 

The proposed method does not require a high-accuracy model of the
environment, but the environment must be decomposed in sufficient
primitives. However, it still requires high-accuracy calibration between
the primitives and the robot. A practical methodology for achieving
this is the topic of ongoing work. An additional topic of ongoing
work is the tracking of a user generated trajectory, which raises
points of discussion such as gain tunning for minimum trajectory tracking
error and latency analysis. The implementation on the real platform
is currently in progress.\bibliographystyle{ieeetr}
\bibliography{folder_safe,additional_bibliography}

\begin{thebibliography}{10}

\bibitem{lum2009raven}
M.~J. Lum, D.~C. Friedman, G.~Sankaranarayanan, H.~King, K.~Fodero,
  R.~Leuschke, B.~Hannaford, J.~Rosen, and M.~N. Sinanan, ``The raven: Design
  and validation of a telesurgery system,'' {\em The International Journal of
  Robotics Research}, vol.~28, no.~9, pp.~1183--1197, 2009.

\bibitem{taylor1999steady}
R.~Taylor, P.~Jensen, L.~Whitcomb, A.~Barnes, R.~Kumar, D.~Stoianovici,
  P.~Gupta, Z.~Wang, E.~Dejuan, and L.~Kavoussi, ``A steady-hand robotic system
  for microsurgical augmentation,'' {\em The International Journal of Robotics
  Research}, vol.~18, no.~12, pp.~1201--1210, 1999.

\bibitem{morita2005microsurgical}
A.~Morita, S.~Sora, M.~Mitsuishi, S.~Warisawa, K.~Suruman, D.~Asai, J.~Arata,
  S.~Baba, H.~Takahashi, R.~Mochizuki, {\em et~al.}, ``Microsurgical robotic
  system for the deep surgical field: development of a prototype and
  feasibility studies in animal and cadaveric models,'' {\em Journal of
  neurosurgery}, vol.~103, no.~2, pp.~320--327, 2005.

\bibitem{UEDA2017110}
H.~Ueda, R.~Suzuki, A.~Nakazawa, Y.~Kurose, M.~M. Marinho, N.~Shono,
  H.~Nakatomi, N.~Saito, E.~Watanabe, A.~Morita, K.~Harada, N.~Sugita, and
  M.~Mitsuishi, ``Toward autonomous collision avoidance for robotic
  neurosurgery in deep and narrow spaces in the brain,'' {\em Procedia CIRP},
  vol.~65, no.~Supplement C, pp.~110 -- 114, 2017.
\newblock 3rd CIRP Conference on BioManufacturing.

\bibitem{li2007spatial}
M.~Li, M.~Ishii, and R.~H. Taylor, ``Spatial motion constraints using virtual
  fixtures generated by anatomy,'' {\em IEEE Transactions on Robotics},
  vol.~23, no.~1, pp.~4--19, 2007.

\bibitem{Siciliano2009}
B.~Siciliano, L.~Sciavicco, L.~Villani, and G.~Oriolo, {\em {Robotics:
  Modelling, Planning and Control}}.
\newblock Advanced Textbooks in Control and Signal Processing, London:
  Springer-Verlag London, 2009.

\bibitem{kapoor2006constrained}
A.~Kapoor, M.~Li, and R.~H. Taylor, ``Constrained control for surgical
  assistant robots.,'' in {\em Robotics and Automation (ICRA), 2006 IEEE
  International Conference on}, pp.~231--236, 2006.

\bibitem{aghakhani2013task}
N.~Aghakhani, M.~Geravand, N.~Shahriari, M.~Vendittelli, and G.~Oriolo, ``Task
  control with remote center of motion constraint for minimally invasive
  robotic surgery,'' in {\em Robotics and Automation (ICRA), 2013 IEEE
  International Conference on}, pp.~5807--5812, IEEE, 2013.

\bibitem{pham2015analysis}
C.~D. Pham, F.~Coutinho, A.~C. Leite, F.~Lizarralde, P.~J. From, and
  R.~Johansson, ``Analysis of a moving remote center of motion for
  robotics-assisted minimally invasive surgery,'' in {\em Intelligent Robots
  and Systems (IROS), 2015 IEEE/RSJ International Conference on},
  pp.~1440--1446, IEEE, 2015.

\bibitem{marinho2016using}
M.~M. Marinho, M.~C. Bernardes, and A.~P. Bo, ``Using general-purpose
  serial-link manipulators for laparoscopic surgery with moving remote center
  of motion,'' {\em Journal of Medical Robotics Research}, vol.~1, no.~04,
  p.~1650007, 2016.

\bibitem{bowyer2014active}
S.~A. Bowyer, B.~L. Davies, and F.~R. y~Baena, ``Active constraints/virtual
  fixtures: A survey,'' {\em IEEE Transactions on Robotics}, vol.~30, no.~1,
  pp.~138--157, 2014.

\bibitem{chen2016virtual}
Z.~Chen, A.~Malpani, P.~Chalasani, A.~Deguet, S.~S. Vedula, P.~Kazanzides, and
  R.~H. Taylor, ``Virtual fixture assistance for needle passing and knot
  tying,'' in {\em Intelligent Robots and Systems (IROS), 2016 IEEE/RSJ
  International Conference on}, pp.~2343--2350, IEEE, 2016.

\bibitem{vitrani2017applying}
M.-A. Vitrani, C.~Poquet, and G.~Morel, ``Applying virtual fixtures to the
  distal end of a minimally invasive surgery instrument,'' {\em IEEE
  Transactions on Robotics}, vol.~33, no.~1, pp.~114--123, 2017.

\bibitem{funda1996constrained}
J.~Funda, R.~H. Taylor, B.~Eldridge, S.~Gomory, and K.~G. Gruben, ``Constrained
  cartesian motion control for teleoperated surgical robots,'' {\em IEEE
  Transactions on Robotics and Automation}, vol.~12, no.~3, pp.~453--465, 1996.

\bibitem{gonccalves2016parsimonious}
V.~M. Gon{\c{c}}alves, P.~Fraisse, A.~Crosnier, and B.~V. Adorno,
  ``Parsimonious kinematic control of highly redundant robots,'' {\em IEEE
  Robotics and Automation Letters}, vol.~1, no.~1, pp.~65--72, 2016.

\bibitem{Quiroz-Omana2017}
J.~J. Quiroz-Oma{\~{n}}a and B.~V. Adorno, ``{Whole-Body Kinematic Control of
  Nonholonomic Mobile Manipulators Using Linear Programming},'' {\em Journal of
  Intelligent {\&} Robotic Systems}, in press.

\bibitem{xia2008integrated}
T.~Xia, C.~Baird, G.~Jallo, K.~Hayes, N.~Nakajima, N.~Hata, and P.~Kazanzides,
  ``An integrated system for planning, navigation and robotic assistance for
  skull base surgery,'' {\em The International Journal of Medical Robotics and
  Computer Assisted Surgery}, vol.~4, no.~4, pp.~321--330, 2008.

\bibitem{prada2009study}
R.~Prada and S.~Payandeh, ``On study of design and implementation of virtual
  fixtures,'' {\em Virtual reality}, vol.~13, no.~2, pp.~117--129, 2009.

\bibitem{Adorno2011e}
B.~V. Adorno, {\em {Two-arm Manipulation: From Manipulators to Enhanced
  Human-Robot Collaboration [Contribution {\`{a}} la manipulation {\`{a}} deux
  bras : des manipulateurs {\`{a}} la collaboration homme-robot]}}.
\newblock {PhD Dissertation}, Universit{\'{e}} Montpellier 2, 2011.

\bibitem{yang1963application}
A.~T. Yang, {\em Application of quaternion algebra and dual numbers to the
  analysis of spatial mechanisms}.
\newblock PhD thesis, Columbia University., 1963.

\bibitem{Perez2004}
A.~Perez and J.~M. McCarthy, ``{Dual Quaternion Synthesis of Constrained
  Robotic Systems},'' {\em Journal of Mechanical Design}, vol.~126, no.~3,
  pp.~425--435, 2004.

\bibitem{adorno2017robot}
B.~V. Adorno, ``Robot kinematic modeling and control based on dual quaternion
  algebra --- {Part I}: Fundamentals,'' 2017.

\bibitem{Selig2005}
J.~M. Selig, {\em {Geometric fundamentals of robotics}}.
\newblock Springer-Verlag New York Inc., 2nd~ed., 2005.

\bibitem{adorno2010dual}
B.~V. Adorno, P.~Fraisse, and S.~Druon, ``Dual position control strategies
  using the cooperative dual task-space framework,'' in {\em Intelligent Robots
  and Systems (IROS), 2010 IEEE/RSJ International Conference on},
  pp.~3955--3960, IEEE, 2010.

\bibitem{Escande2014}
A.~Escande, N.~Mansard, and P.-B. Wieber, ``{Hierarchical quadratic
  programming: Fast online humanoid-robot motion generation},'' {\em The
  International Journal of Robotics Research}, vol.~33, pp.~1006--1028, may
  2014.

\end{thebibliography}
 
\end{document}